\documentclass[11pt]{article}
\usepackage{iftex}
\ifPDFTeX
  \usepackage[utf8]{inputenc}
  \usepackage[T1]{fontenc}
  \usepackage{tgtermes}% Times-like text font (metric-compatible with Liberation Serif)
\else
  \usepackage{fontspec}
\fi
\usepackage{microtype}
\usepackage[top=1in,bottom=1in,left=1in,right=1in]{geometry}
\usepackage{amsmath,amsthm,amssymb}
\usepackage{booktabs}
\usepackage{multirow}
\usepackage{graphicx}
\usepackage{subcaption}
\usepackage{xcolor}
\usepackage{hyperref}
\usepackage{cleveref}
\usepackage{enumitem}
\usepackage{url}
\usepackage{array}
\usepackage{placeins}
\usepackage{listings}
\usepackage[numbers]{natbib} 
\definecolor{pipelineblue}{RGB}{52,100,178}
\definecolor{codegray}{RGB}{245,245,245}

\hypersetup{
  pdftitle={AraMS-28k: The Largest Publicly Released Line-Level Dataset of
            Historical Arabic Manuscripts with Margin and Insertion-Anchor
            Annotations},
  pdfauthor={Mohamed Guechaoui, Mohamed Diaa Zellagui, Souleyman Chaib, Sahraoui Dhelim},
  colorlinks=true,
  linkcolor=pipelineblue,
  citecolor=pipelineblue,
  urlcolor=pipelineblue
}
\title{\textbf{AraMS-28k: The Largest Publicly Released Line-Level Dataset of
  Historical Arabic Manuscripts with Margin and Insertion-Anchor
  Annotations}}
\author{%
  Mohamed Guechaoui$^1$ \and Mohamed Diaa Zellagui$^1$
  \and Souleyman Chaib$^1$ \and Sahraoui Dhelim$^1$
  \\[2pt]
  $^1$Higher School of Computer Science (ESI-SBA), Sidi Bel Abbes, Algeria
}
\date{}
\begin{document}
\maketitle

%%% ── ABSTRACT ──────────────────────────────────────────────────────────────

\begin{abstract}

We introduce \textbf{AraMS-28k}, the largest publicly released line-level
dataset of genuine historical Arabic manuscripts, comprising 14 books,
3,043 pages, and 28,600 annotated text lines (27,971 main-text, 629
margin). Thirteen books are hand-copied manuscripts spanning three
script traditions -- Naskh, Ruq\textquoteleft ah, and Maghrebi -- and one is
a lithographed printed edition included to broaden format diversity.
Each line is labelled as main-text or margin, and margin lines that
have an unambiguous attachment point in the main text are further
annotated with an \emph{insertion anchor}, recovering the manuscript's
true non-linear reading order at line-level granularity -- to our
knowledge the first such annotation released for a historical Arabic
manuscript corpus. Because reference transcriptions are fully
vocalised while manuscript hands are typically undiacritised, we
release both the raw diacritised transcription and a
diacritic-normalised counterpart for every line. The dataset was
constructed with RefLAM~\cite{guechaoui2026reflam}, a reference-grounded annotation pipeline that
aligns multimodal-LLM OCR against independently sourced clean
transcriptions and routes every line through human review, combining
automatic verification with expert oversight. We describe the
construction and quality-control process, present the annotation
schema, report dataset statistics at both the corpus and per-book
level, and provide baseline HTR results using Kraken and HATFormer,
including a cross-script generalisation gradient from in-distribution
pages to a fully unseen books. AraMS-28k is released with page
images, line-level annotations, and fixed train/val/test splits under
CC~BY-NC-SA~4.0 to support reproducible research on Arabic manuscript
recognition, layout analysis, and reading-order recovery.
\end{abstract}
%%% ── 1  INTRODUCTION ────────────────────────────────────────────────────────
\section{Introduction}
\label{sec:intro}
Handwritten text recognition (HTR) for historical Arabic manuscripts
lags behind Latin-script HTR largely because of a shortage of large,
line-level corpora drawn from genuine manuscripts rather than modern
handwriting samples or synthetic renderings. Existing public resources
for historical Arabic HTR are limited in scale, restricted to modern
handwriting rather than genuine manuscript material, or -- even when
drawn from real manuscripts -- do not record \emph{where} a given
margin annotation belongs within the main-text reading flow;
\cref{sec:related} compares these resources in detail. This combination
of gaps motivates AraMS-28k.
Margin content in historical Arabic manuscripts is rarely incidental.
A marginal note is usually anchored to a specific point in the main
text: it is a correction, an alternate reading, or a gloss meant to be
read in place, not as an appendix to the page. Recognising a margin
region, or even placing it correctly in a coarse page-level reading
order, is not the same as recording \emph{which line} of the main text
it belongs next to. Recovering this fine-grained, non-linear reading
order requires an annotation that goes beyond a bounding box, a line
transcription, or a region-level layout label -- it requires recording,
at line-level granularity, where each margin line is intended to be
read.
\textbf{AraMS-28k} is built to supply this annotation across a large
corpus of genuine manuscripts. Every line in the corpus is labelled as
main-text or margin, and each margin line is further annotated with an
\emph{insertion anchor} -- the index of the main-text line after which
it logically inserts -- wherever the manuscript provides an unambiguous
attachment point; roughly 30\% of margin lines meet this criterion
(\cref{sec:schema}), while the remainder are retained with a null
anchor rather than a forced, unverifiable guess. To our knowledge, no
prior publicly released Arabic manuscript corpus provides this
line-level reading-order annotation at all: existing resources may
identify marginal text as a distinct region, or in some cases order
regions at the page level, but none record where a specific margin
line attaches within the main text. Separately, because the reference
transcriptions we align against are drawn from fully-vocalised
scholarly editions while the manuscript hands themselves are typically
undiacritised, we also release a diacritic-normalised transcription
alongside the raw one for every line (\cref{sec:schema}) -- a mismatch
between reference and image that, as far as we are aware, existing
Arabic manuscript datasets do not document or address explicitly.
Combined with 28,600 line-level transcriptions across three script
traditions, this makes AraMS-28k usable both as an HTR training corpus
and as a resource for manuscript layout analysis and reading-order
recovery.
\paragraph{Contributions.}
\begin{enumerate}[leftmargin=*,nosep]
  \item \textbf{AraMS-28k}: the largest publicly released line-level
    dataset of genuine historical Arabic manuscripts -- 14 books, 3,043
    pages, and 28,600 line-level annotations spanning three hand-copied
    script traditions (Naskh, Ruq\textquoteleft ah, Maghrebi) and one
    lithographed volume, released under CC~BY-NC-SA~4.0 with fixed
    train/val/test splits (\cref{sec:dataset}).
  \item \textbf{Insertion-anchor annotation}: the first publicly
    released line-level annotation recovering the non-linear
    main/margin reading order in a historical Arabic manuscript
    corpus, assigned wherever a margin line has an unambiguous
    attachment point in the main text (\cref{sec:schema}).
  \item \textbf{A two-phase construction and quality-control process}
    (\cref{sec:construction}) combining automated reference-grounded
    OCR alignment, verified against independently sourced reference
    transcriptions, with human expert review.
  \item \textbf{Baseline HTR results and a cross-script generalisation
    gradient} (\cref{sec:baselines}) for two recognition architectures,
    establishing AraMS-28k as a benchmark for future work.
  \item \textbf{Dual raw/normalised transcriptions}: every line carries
    both the fully diacritised reference form and a diacritic-normalised
    form matched to what is visually present in the manuscript hand
    (\cref{sec:schema}), so downstream users can pick the target that
    matches their task.
\end{enumerate}
\FloatBarrier
%%% ── 2  RELATED WORK ────────────────────────────────────────────────────────
\section{Related Arabic HTR Datasets}
\label{sec:related}
RASM2018~\cite{clausner2018rasm} provides line-level transcriptions of
historical Arabic scientific manuscripts but is limited to
$\approx$120 pages. While it includes region-level layout labels, its
annotation does not link individual margin lines to a specific point
in the main-text reading order. RASAM~\cite{vidalgorene2021rasam} is
the first corpus dedicated to the Maghrebi script family
($\approx$300 pages, 7,540 lines, line-level transcriptions) and the
closest existing resource to our Maghrebi coverage; it is, however, an
order of magnitude smaller than AraMS-28k and records no margin
insertion-anchor. KHATT~\cite{mahmoud2014khatt} is considerably larger
(4,000 pages, 13,435 lines) but consists of modern handwriting samples
collected for the purpose of dataset construction rather than pages
from historical manuscripts, so it does not capture the degradation,
ligature variation, or marginalia found in genuine manuscript material,
OpenITI MAKHZAN \cite{allen2026openiti} is the broadest of these resources in linguistic
scope (1,497 pages, 822 Arabic, across seven Arabic-script languages),
manually segmented and transcribed, but like the other resources above
does not link margin lines to a specific point in the main-text reading
order.Muharaf~\cite{saeed2024muharaf} is the largest existing
genuine-manuscript corpus and our closest comparator: it is drawn from
real manuscripts and includes diverse scripts, and -- as with other
resources in this comparison -- it may mark marginal text as a distinct
layout region, but its annotation does not, to our knowledge, encode
\emph{where} in the main-text reading flow a given margin line belongs.
Muharaf's complete corpus totals $\approx$1,644 pages and 36,311 lines,
but only 1,216 pages (24,495 lines) have been publicly released; we
compare against this public subset in Table~\ref{tab:comparison}, since
it is the relevant baseline for a claim about publicly available
corpora. Among other historical resources,
VML-HD~\cite{kassis2017vmlhd} annotates $\approx$680 pages at the
sub-word level and BADAM~\cite{kiessling2019badam} provides baseline
(rather than transcription) annotations for $\approx$400 pages; both
target tasks other than line-level HTR.
KITAB-Bench~\cite{heakl2025kitab} and SARD~\cite{saeed2025sard} extend
Arabic OCR evaluation to a wider range of document types; KITAB-Bench
includes a small set of historical manuscript samples (HistoryAr and
HistoricalBooks) but does not annotate line-level insertion anchors,
while SARD is fully synthetic. Neither includes manuscript-specific
layout phenomena such as marginalia. We restrict this comparison to
corpora targeting Arabic manuscript or handwriting recognition, since
these are the resources a user of AraMS-28k would realistically
consider as alternatives.
Outside Arabic, IAM~\cite{marti2002iam} and RIMES~\cite{grosicki2009rimes}
are foundational handwriting corpora for English and French
respectively, and both have shaped how line-level HTR datasets are
structured and evaluated. Neither, however, was built around the
two-zone main/margin layout that characterises historical Arabic
manuscripts, since the documents they draw from do not exhibit this
structure. Table~\ref{tab:comparison} situates AraMS-28k against the
Arabic-language corpora above: it is the largest \emph{publicly
released} corpus by line count among genuine historical handwritten
manuscript corpora --Muharaf's full corpus is larger in total, with a restricted subset held under proprietary license -- and, to our knowledge, the only one that links
individual margin lines to a specific point in the main-text reading
order via a line-level insertion anchor. We deliberately restrict this
claim to publicly released corpora throughout the paper; we make no
claim of being the largest Arabic manuscript resource in an absolute
sense, since larger privately held corpora may exist.
\begin{table}[htbp]
  \centering
  \caption{Comparison with existing public Arabic manuscript datasets.
  Figures for Muharaf reflect its publicly released subset only (its
  full corpus totals 1,644 pages / 36,311 lines). ``Insertion anchor''
  indicates whether margin lines are linked to a specific point in the
  main-text reading order -- not merely whether margin text is identified
  as a distinct region.}
  \label{tab:comparison}
  \small
  \renewcommand{\arraystretch}{1.15}
  \begin{tabular}{@{}lrrc@{}}
    \toprule
    Dataset & Pages & Lines & Insertion anchor \\
    \midrule
    RASM2018~\cite{clausner2018rasm}    &  120  &  2,613 & No \\
    RASAM~\cite{vidalgorene2021rasam}   &  300  &  7,540 & No \\
    KHATT~\cite{mahmoud2014khatt}       & 4,000 & 13,435 & No \\
    Muharaf-public~\cite{saeed2024muharaf}     & 1,216 & 24,495 & No \\
    \textbf{AraMS-28k (ours)}           & \textbf{3,043} & \textbf{28,600} & \textbf{Yes} \\
    \bottomrule
  \end{tabular}
\end{table}
\FloatBarrier
%%% ── 3  DATASET CONSTRUCTION ─────────────────────────────────────────────────
\section{Dataset Construction}
\label{sec:construction}
\subsection{Source Material}
\begin{figure*}[tbp]
  \centering
  \begin{subfigure}[t]{0.31\textwidth}
    \includegraphics[width=\linewidth]{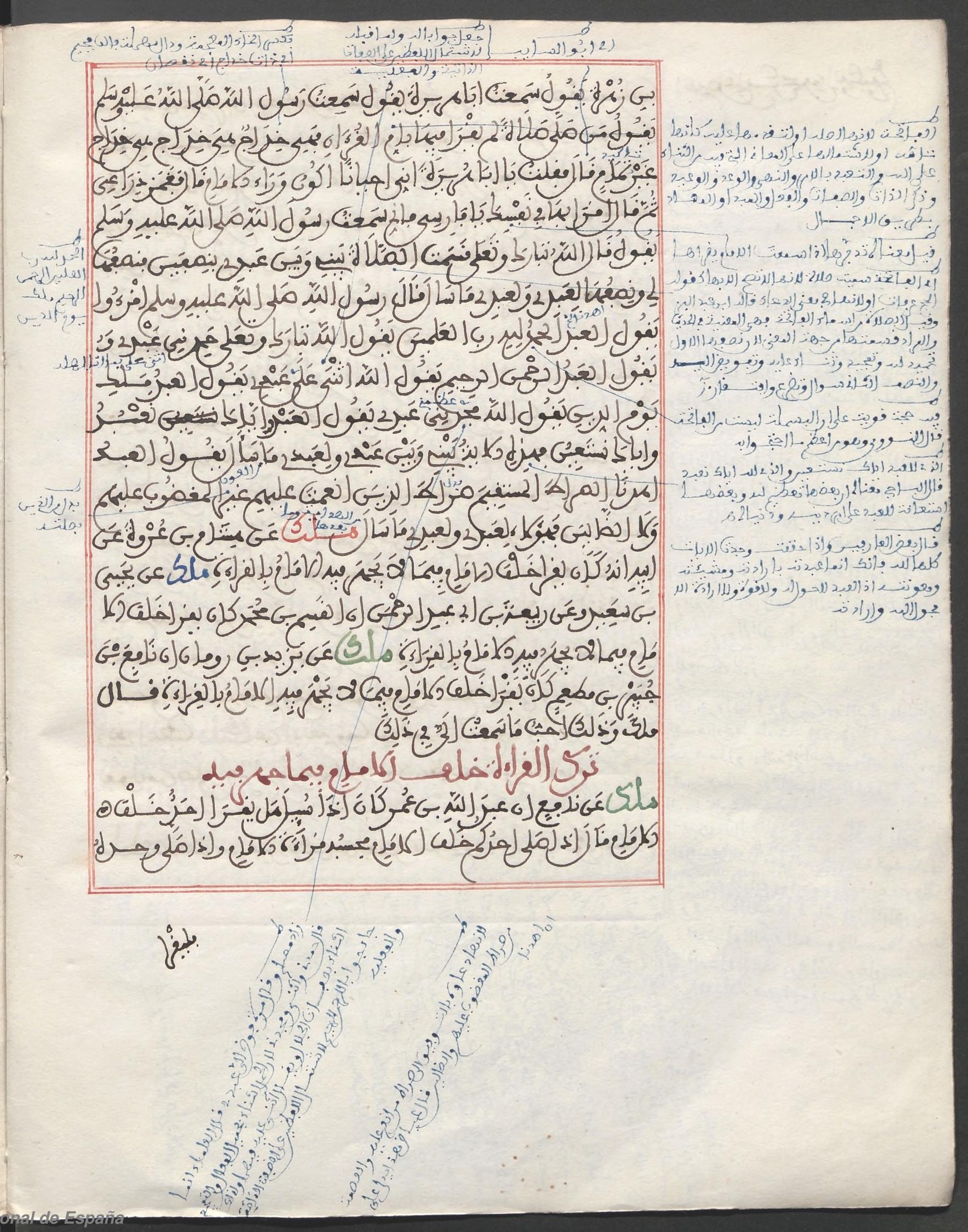}
    \caption{Maghrebi.}
  \end{subfigure}
  \hfill
  \begin{subfigure}[t]{0.31\textwidth}
    \includegraphics[width=\linewidth]{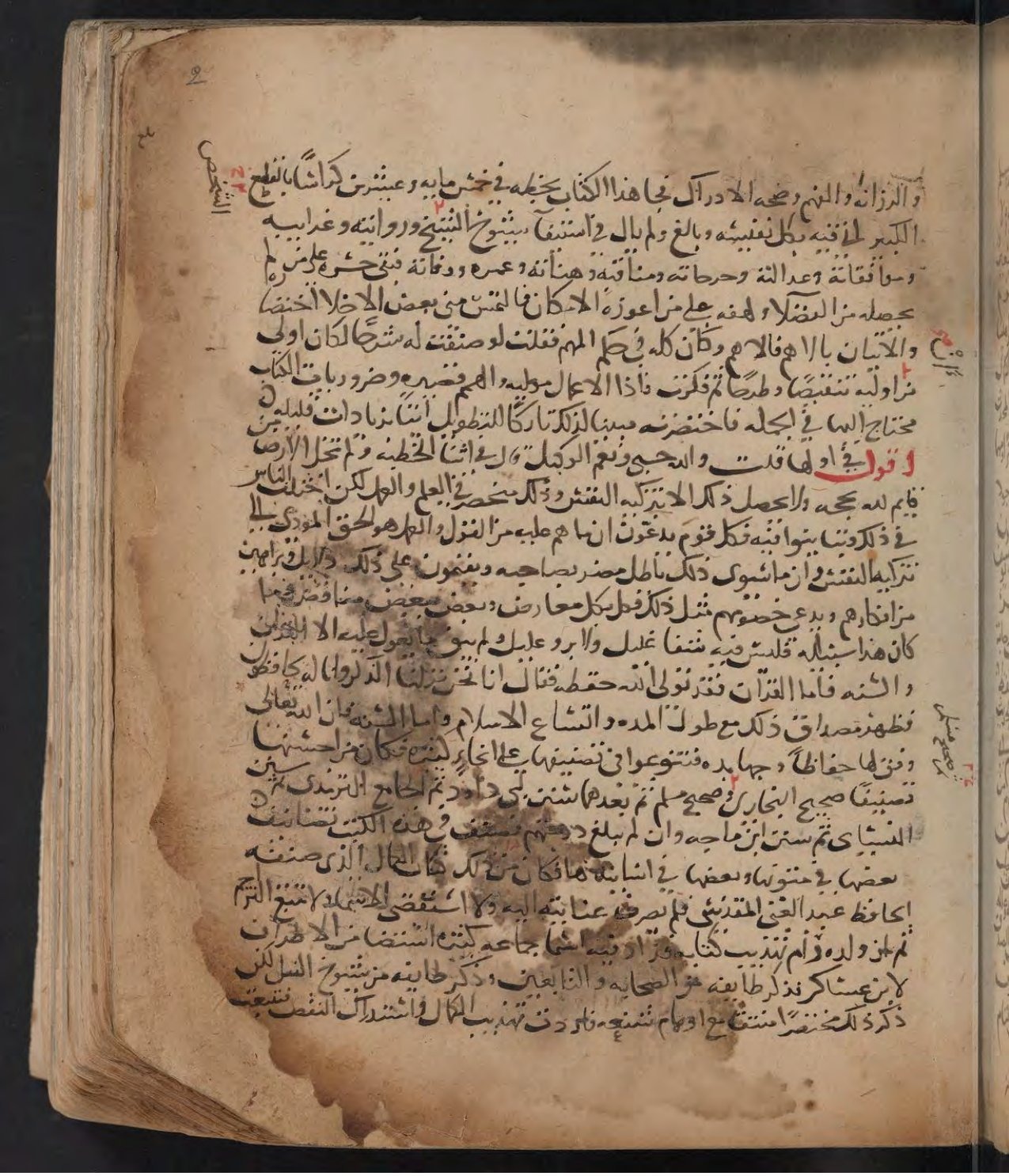}
    \caption{Naskh.}
  \end{subfigure}
  \hfill
  \begin{subfigure}[t]{0.31\textwidth}
    \includegraphics[width=\linewidth]{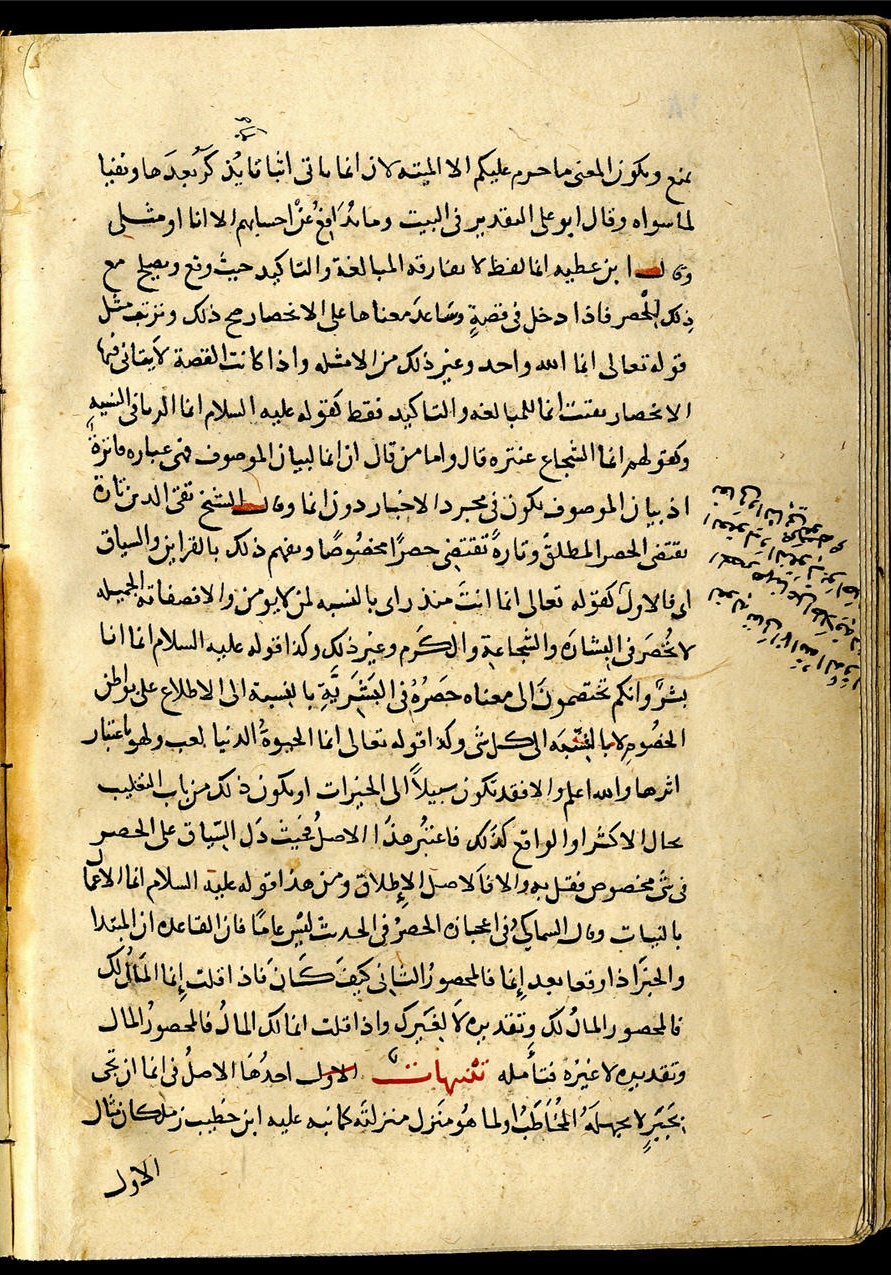}
    \caption{Ruq\textquoteleft ah.}
  \end{subfigure}
  \caption{Script diversity in AraMS-28k. Sample pages illustrate the
    range of hands, ink density, and marginal annotation present across
    the three hand-copied script traditions; the lithographed volume
    (\texttt{book\_10}) is a distinct production format and is not
    pictured here.}
  \label{fig:script_samples}
\end{figure*}

AraMS-28k is drawn from 14 historical Arabic manuscript books spanning
classical medicine, Islamic jurisprudence, Peripatetic philosophy, and
theology. 13 of these are hand-copied manuscripts; one
(\texttt{book\_10}, Table~\ref{tab:perbook}) is a lithographed printed
edition -- a historical Arabic printing technique that reproduces a
scribe's handwriting via a lithographic stone plate rather than
movable type -- and is included to diversify page layout and production
format beyond purely hand-copied material. Source books were selected
to maximise diversity along four axes: script style (Naskh,
Ruq\textquoteleft ah, Maghrebi), diacritisation density, scan quality,
and layout complexity, including the presence and density of marginal
annotation (Figure~\ref{fig:script_samples}). Page images were obtained from publicly accessible online manuscript
repositories (e.g., alukah.net); the per-book source, access terms, and redistribution
basis are recorded in the release manifest
(\texttt{manifest.json}), and only scans whose terms permit
redistribution under the release licence are included. For each book, a
clean ground-truth transcription was obtained either from existing
digital Arabic text repositories and one of the books (book\_27) produced by OCR over a scanned
printed critical edition of the same text, then used as the reference
against which manuscript pages were aligned.
\subsection{Annotation Pipeline}
\begin{figure}[htbp]
  \centering
  \includegraphics[width=0.78\columnwidth,height=0.5\textheight,keepaspectratio]{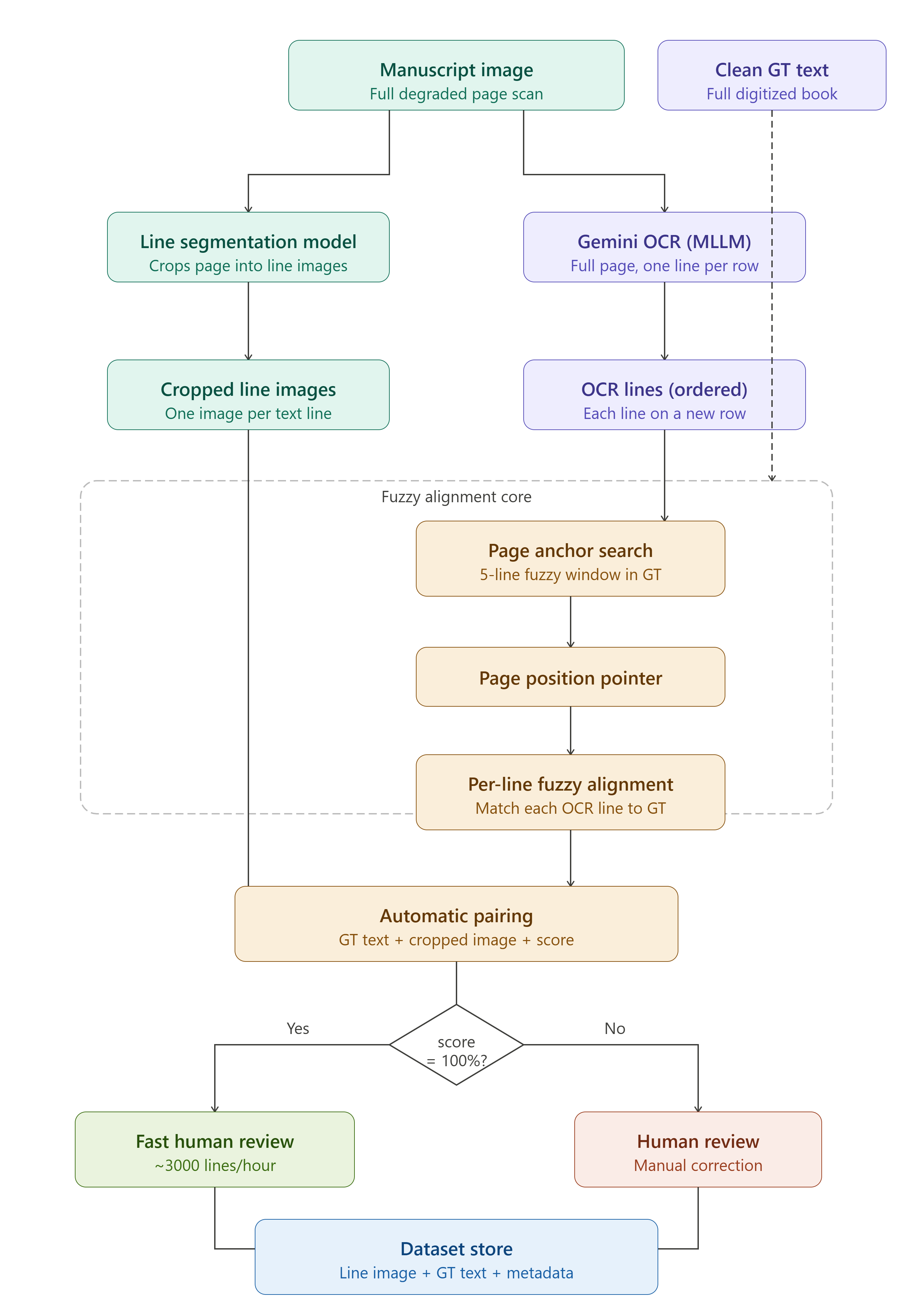}
  \caption{Overview of the RefLAM construction pipeline used to build
    AraMS-28k: line segmentation and multimodal-LLM OCR feed a
    reference-alignment step that routes every line to human review.
    Full detail is given in~\cite{guechaoui2026reflam}.}
  \label{fig:pipeline_overview}
\end{figure}

Pages were annotated with \textbf{RefLAM}, a reference-grounded pipeline
built for this and related corpora. At a high level, RefLAM (1) segments
each page into individual text lines with a segmentation model trained
on Muharaf~\cite{bors2024seg}, (2) transcribes each page with a
multimodal LLM, separating main-text from marginal content, and (3)
aligns the resulting hypothesis against the clean reference
transcription, producing a per-line confidence score and flagging
low-confidence lines for closer inspection
(Figure~\ref{fig:pipeline_overview}). Every line, regardless of its
confidence score, was reviewed by a human annotator before release;
lines with a perfect alignment score were fast-reviewed for gross
segmentation errors, while lower-confidence lines received full manual
correction. We refer to the RefLAM paper~\cite{guechaoui2026reflam} for the
alignment algorithm and its correctness properties; here we treat it
only as an already-validated construction tool.
\subsection{Quality Control and Validation}
\label{sec:qc}
Every line in AraMS-28k passes through a verification step before
release: no automatically generated annotation is accepted without
human sign-off. Construction proceeded in two phases that differ in
review depth, chosen to trade off annotation depth against corpus
scale.
In the \emph{page-validated} (PV) phase, covering 7 books (548 pages,
11,438 main lines), two independent reviewers inspected every line on
every page regardless of the automatic alignment confidence score,
providing full manual verification of both segmentation and
transcription. In the \emph{line-validated} (LV) phase, covering the
remaining 7 books (2,495 pages), the automatic reference-alignment step
is used as a formal acceptance filter: only main-text lines that
achieve a perfect alignment score against the independently sourced
reference transcription are retained, and any line falling short of
this threshold is excluded from the release rather than manually
corrected, so that scale is not purchased at the cost of admitting
unverified text. Because the
reference transcription is drawn from an independently produced
critical edition rather than derived from the manuscript image itself,
a perfect alignment is a meaningful correctness signal rather than a
self-consistency check: it indicates that an OCR hypothesis produced
from the image independently reproduces text already known, from an
external source, to be correct.
This two-tier design has a direct, disclosed consequence for downstream
use: LV statistics reflect a filtered, agreement-verified subset that
is plausibly biased toward cleaner scans and more regular hands, while
PV statistics reflect a page's full line content, filtered only by
human review rather than by automatic agreement. We report both
subsets separately throughout (\cref{sec:dataset} and
Appendix~\ref{app:perbook}) rather than merging them, so that users can
select the guarantee level appropriate to their task: full manual
verification (PV), or scale combined with automatic reference agreement
(LV).
\subsection{Construction Timeline}
\label{sec:timeline}
Most  of the  corpus --  14 books, 3,043 pages, and 28,600 lines -- was
constructed in approximately one week. This pace was driven primarily
by the line-validated (LV) phase (\cref{sec:qc}): because LV admits a
line only on perfect automatic alignment against an independently
sourced reference, human effort there is limited to spot-checking
accepted lines and reviewing page-level segmentation, rather than
transcribing or correcting each line by hand. The page-validated (PV)
phase, which requires full manual verification of every line by two
reviewers, is correspondingly slower per page and accounts for a
proportionally larger share of the total annotation time despite
covering only 548 of the corpus's 3,043 pages. We view this as
evidence that a reference-grounded, LLM-assisted pipeline can make
large-scale historical manuscript annotation tractable at a fraction
of the wall-clock cost of purely manual transcription; a controlled
timing study of the review workflow, including a measured
manual-annotation baseline, is reported in the RefLAM
paper~\cite{guechaoui2026reflam}.
\FloatBarrier
%%% ── 4  ANNOTATION SCHEMA ───────────────────────────────────────────────────
\section{Annotation Schema}
\label{sec:schema}
\begin{figure}[htbp]
  \centering
  \includegraphics[width=0.62\columnwidth,height=0.52\textheight,keepaspectratio]{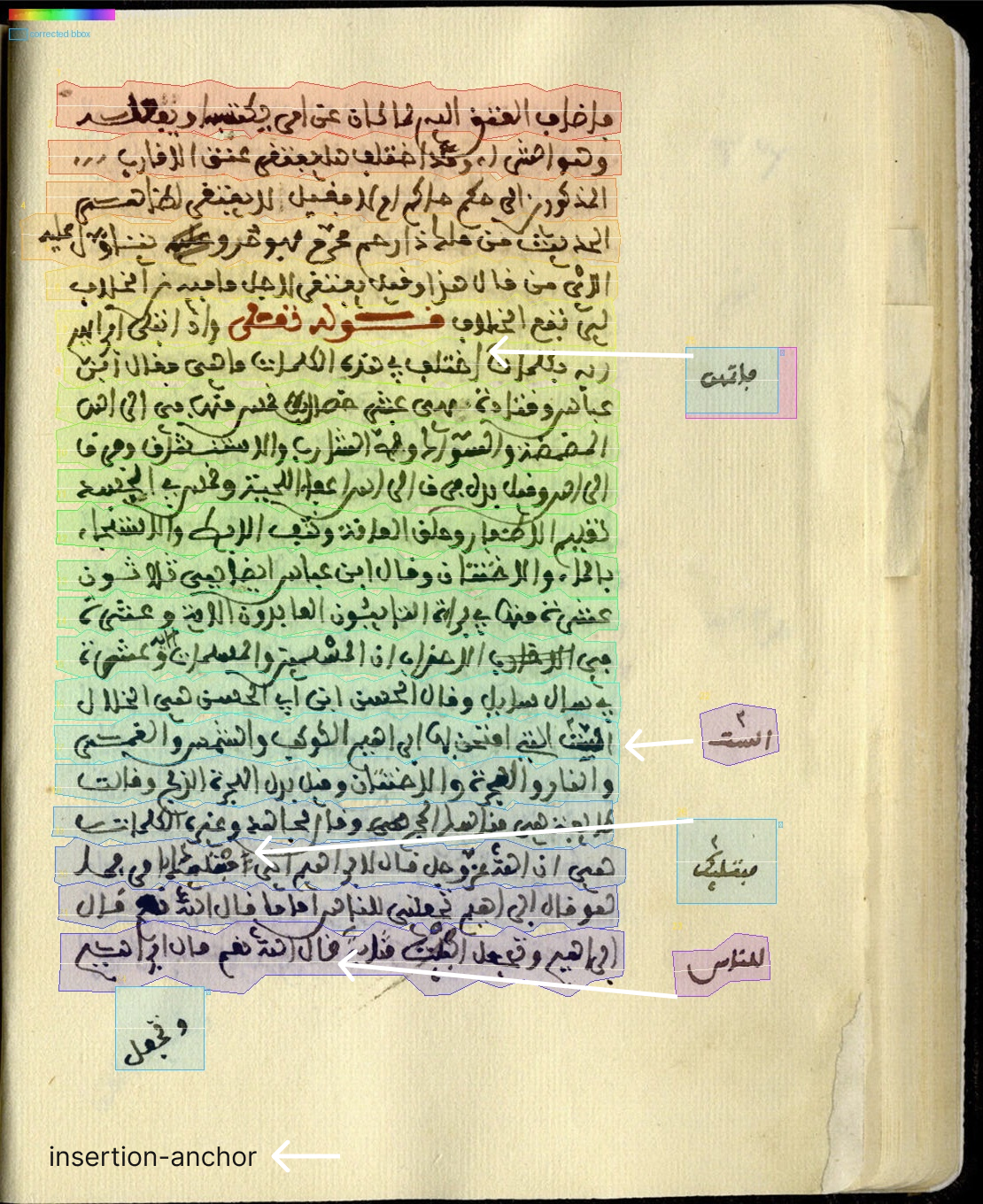}
  \caption{Insertion-anchor annotation. Coloured regions mark segmented
    main and margin lines; the arrow indicates the main-text line after
    which the highlighted margin line logically inserts.}
  \label{fig:insertion_anchor_example}
\end{figure}

Each annotated line record in our dataset is structured into five logical groups:
(i) basic identifiers and page metadata;
(ii) spatial geometry (baseline, polygon, or bounding box);
(iii) transcriptions and alignment confidence;
(iv) margin-specific anchoring metadata; and
(v) human review flags.
Table~\ref{tab:schema} provides a complete overview of all fields.
\begin{table}[htbp]
\centering
\small
\begin{tabular}{p{2.8cm}p{1.8cm}p{7.5cm}}
\toprule
\textbf{Field} & \textbf{Type} & \textbf{Description} \\
\midrule
\multicolumn{3}{l}{\textit{Top-level identifiers and metadata}} \\
\texttt{line\_uid} & string & Globally unique line ID, e.g., \texttt{book\_03\_page\_076\_L0000} \\
\texttt{book\_id} & string & Book identifier, e.g., \texttt{book\_03} \\
\texttt{page\_id} & string & Page identifier, e.g., \texttt{book\_03\_page\_076} \\
\texttt{page\_image} & string & Relative path to the full-page image \\
\texttt{line\_idx} & integer & Zero-based line index on the page (top-to-bottom) \\
\texttt{line\_type} & string & \texttt{main} or \texttt{margin} \\
\texttt{split} & string & Dataset split: \texttt{train}, \texttt{val}, or \texttt{test} \\
\midrule
\multicolumn{3}{l}{\textit{text (Transcriptions)}} \\
\texttt{gt\_raw} & string & Clean reference transcription from the scholarly edition \\
\texttt{gemini\_raw} & string & Raw MLLM OCR hypothesis before correction \\
\texttt{gt\_normalized} & string & Diacritic-normalised form of \texttt{gt\_raw}; recommended HTR training target \\
\texttt{confidence} & number & Alignment confidence in [0, 100]; 100 guarantees character-for-character identity \\
\midrule
\multicolumn{3}{l}{\textit{geometry (Spatial layout)}} \\
\texttt{baseline} & array / null & Ordered polyline tracing the text baseline (the line of writing itself) \\
\texttt{boundary\_polygon} & array / null & Ordered polygon vertices \texttt{[[x,y], ...]} outlining the line's surrounding region \\
\texttt{bounding\_box} & object / null & Axis-aligned bounding box \texttt{\{x, y, w, h\}} \\
\midrule
\multicolumn{3}{l}{\textit{margin\_anchor (Margin metadata, null for main lines)}} \\
\texttt{before} & string / null & Main-text words immediately following the insertion point \\
\texttt{after} & string / null & Main-text words immediately preceding the insertion point \\
\texttt{line} & integer / null &Index of the main-text line the margin note is anchored to; the exact
insertion point (including mid-line insertions) is given by
\texttt{before}/\texttt{after}. \\
\texttt{rotation} & integer / null & Coarse orientation in degrees (e.g., \texttt{90}, \texttt{-90}); \texttt{null} denotes horizontal \\
\midrule
\multicolumn{3}{l}{\textit{review (Human quality control)}} \\
\texttt{edited} & boolean & Whether the transcription was manually edited \\
\texttt{validated} & boolean & Whether the line passed human validation \\
\texttt{deleted} & boolean & Whether the line is marked for deletion \\
\texttt{page\_reviewed} & boolean & Whether the entire containing page was reviewed \\
\bottomrule
\end{tabular}
\caption{Complete annotation schema for the dataset. Every key is
present in every record; geometry and margin fields take the value
\texttt{null} when not applicable.}
\label{tab:schema}
\end{table}
The following minimal record illustrates a typical \texttt{main} line. Note that the \texttt{geometry} object is present but set to \texttt{null} for brevity, and \texttt{margin\_anchor} is explicitly set to \texttt{null}.
\begin{lstlisting}
{
  "line_uid": "book_03_page_001_L0000",
  "book_id": "book_03",
  "page_id": "book_03_page_001",
  "page_image": "images/book_03/book_03_page_001.jpg",
  "line_idx": 0,
  "line_type": "main",
  "text": {
    "gt_raw": "...",
    "gemini_raw": "...",
    "confidence": 100,
    "gt_normalized": "...",

    
  },
  "geometry": {
    "baseline": null,
    "boundary_polygon": null,
    "bounding_box": null
  },
  "margin_anchor": null,
  "review": {
    "edited": false,
    "validated": true,
    "deleted": false,
    "page_reviewed": true
  },
  "split": "test"
}
\end{lstlisting}
For \texttt{margin} lines, the \texttt{margin\_anchor} object is populated instead of being \texttt{null}. For example:
\begin{lstlisting}
"margin_anchor": {
  "before": "<main-text words  following the insertion point>",
  "after": "<main-text words  preceding the insertion point>",
  "line": 5,
  "rotation": 90 , 
}
\end{lstlisting}
\paragraph{Diacritics and normalisation.} Reference transcriptions are
drawn from fully-vocalised scholarly editions and therefore contain
diacritical marks (\emph{tashkeel}), the kashida (\emph{tatweel}), and
specific alef/hamza letter forms that are frequently absent, or applied
only inconsistently, in the manuscript hand itself. Training a
recognition model directly against \texttt{gt\_raw} therefore implicitly
asks it to predict symbols with no visual evidence in the input image --
a mismatch that is easy to overlook and, left undocumented, can silently
inflate character error rate or mislead a model toward hallucinating
diacritics. To make this an explicit, usable choice rather than a tacit
one, every line carries a second transcription field,
\texttt{gt\_normalized}: the output of the same normalisation operator
(Definition~1 in~\cite{guechaoui2026reflam}) that RefLAM uses internally to align
OCR against the reference, applied here to the reference text itself and
persisted rather than discarded after alignment. Concretely,
\texttt{gt\_normalized} strips harakat and tatweel, merges alef and
alef-maq\d{s}\=ura/hamza variants, and collapses whitespace, yielding a
transcription that matches what is visually present in an undiacritised
manuscript hand. \textbf{We recommend \texttt{gt\_normalized} as the
training target for page-transcription models}, and reserve
\texttt{gt\_raw} for diacritisation-restoration research or use cases
where the fully vocalised scholarly reading is itself the desired
output.
\paragraph{Insertion anchors: a precision-first annotation.} Historical
Arabic manuscript pages are frequently non-linear: a margin note is
physically outside the main text block but is intended to be read at a
specific point within it. AraMS-28k records this explicitly through the
\texttt{margin\_anchor} record, whose \texttt{line} field gives the
index of the main-text line after which a given margin line logically
inserts (Figure~\ref{fig:insertion_anchor_example}). Anchors are
assigned only where the margin content has an unambiguous, verifiable
attachment point in the main text; under this criterion, human
annotators assign a confident anchor to 189 of the 629 margin lines in
the corpus ($\approx$30\%). The remainder -- ownership stamps, later
commentary, or placements too ambiguous to anchor confidently -- are
retained with a null anchor rather than a forced guess.
This is a deliberate precision-over-coverage design choice, not a
limitation of the annotation process: an incorrectly forced anchor
would misrepresent the manuscript's true reading order and could
silently corrupt any downstream evaluation of reading-order recovery,
whereas a null anchor correctly and explicitly signals ``no confident
attachment point exists.'' Because the field is ternary rather than
binary at the corpus level -- present, confidently null, or absent from
non-margin lines entirely -- downstream users can trivially filter to
the 189 confidently anchored lines when a clean, high-precision
reading-order signal is required, or use the full margin set together
with the null-anchor flag when studying margin content more broadly
(e.g., margin detection or classification of marginalia type).
\FloatBarrier
%%% ── 5  DATASET STATISTICS ──────────────────────────────────────────────────
\section{Dataset Statistics}
\label{sec:dataset}
AraMS-28k contains 3,043 manuscript pages and 28,600 annotated text
lines (27,971 main-text, 629 margin) across 14 books. Thirteen books
are hand-copied manuscripts spanning three script traditions (Naskh,
Ruq\textquoteleft ah, Maghrebi); the remaining book is a lithographed
printed edition (\texttt{book\_10}, Table~\ref{tab:perbook}), included
to diversify page layout and print format beyond purely hand-copied
material. Table~\ref{tab:dataset_summary} summarises the released
corpus; Table~\ref{tab:perbook} in Appendix~\ref{app:perbook} breaks
these totals down per book, separately for the page-validated (PV) and
line-validated (LV) subsets described in \cref{sec:qc}.
\begin{table}[htbp]
  \centering
  \caption{AraMS-28k summary statistics.}
  \label{tab:dataset_summary}
  \small
  \renewcommand{\arraystretch}{1.15}
  \begin{tabular}{@{}p{4.4cm}p{3cm}@{}}
    \toprule
    \textbf{Property} & \textbf{Value} \\
    \midrule
    Books & 14 \\
    Pages & 3,043 \\
    \quad Page-validated (PV) & 548 \\
    \quad Line-validated (LV) & 2,495 \\
    Main lines & 27,971 \\
    Margin lines & 629 \\
    Scripts & Naskh, Ruq\textquoteleft ah, \\
            & Maghrebi (+1 lithographed vol.) \\
    Confidently anchored margin lines & 189 / 629 ($\approx$30\%) \\
    Train / val / test split & 9 / 2 / 3 books \\
    License & CC BY-NC-SA 4.0 \\
    \bottomrule
  \end{tabular}
\end{table}
Splits are formed at the book level rather than the page or line level,
so that no manuscript appears in more than one split. The training set
(9 books, 19,739 lines) mixes PV and LV books across all three
hand-copied script traditions; the validation set (2 books, 1,486
lines) is page-validated Naskh material; and the held-out test set
(3 books, 6,746 lines) is exclusively page-validated, with one book per
hand-copied script, so that baseline results in \cref{sec:baselines}
can be reported per script. Together the three splits account for all
27,971 main-text lines. Per-book line counts, average lines per page,
and average words per line are given in Appendix~\ref{app:perbook}.
\paragraph{Supported tasks.} AraMS-28k is a benchmark for HTR
(\cref{sec:baselines}). The released geometry of every line is
human-verified, and the 548 PV pages carry every line on the page, so
the PV subset additionally supports text-line detection and
segmentation research; LV pages contain only the reference-verified
subset of their lines (\cref{sec:qc}) and are therefore suitable for
recognition, but not as exhaustive detection ground truth. The
main/margin labels and insertion anchors further enable
layout-analysis and reading-order-recovery research; for these two
tasks we release the annotations but define no evaluation protocol or
baseline, which we leave to future work. Finally, the release flags a
curated subset of 177 naturally degraded lines from \texttt{book\_09}
(\texttt{real\_damage\_lines.txt}), selected for severe native
degradation such as fading, staining, and bleed-through; because
\texttt{book\_09} lies in the held-out test split, this subset provides
a ready-made out-of-distribution evaluation set for document-restoration
and recognition-robustness research.
\FloatBarrier
%%% ── 6  BASELINE EXPERIMENTS ────────────────────────────────────────────────
\section{Baseline Recognition Results}
\label{sec:baselines}
\paragraph{Setup.} We finetune two Muharaf-pretrained recognisers --
Kraken~\cite{kiessling2019kraken}, using the publicly released
Muharaf-trained checkpoint~\cite{bors2024rec}, and
HATFormer~\cite{chan2025hatformer} -- on the AraMS-28k training split
and evaluate on the held-out test split, which contains one book per
hand-copied script. All recognition experiments use main-text lines
only (margin lines are excluded), the fixed book-levels splits
distributed with the release, and \texttt{gt\_normalized} as the
transcription target: computing CER against the diacritised
\texttt{gt\_raw} would conflate recognition error with unrecoverable
diacritisation, since the model receives no visual signal for
diacritics absent from the manuscript hand. Both models are finetuned
with standard recipes from their respective pretrained checkpoints;
full hyperparameters, finetuning code, and final-evaluation decoding
settings are given in Appendix~\ref{app:finetuning-details} and
distributed in the project repository released alongside the
dataset, so every figure in this section can be regenerated from the
released splits. Table~\ref{tab:baseline} reports character error
rate (CER) per test book and overall.

\begin{table}[htbp]
  \centering
  \caption{CER after finetuning Muharaf-pretrained models on AraMS-28k.}
  \label{tab:baseline}
  \small
  \renewcommand{\arraystretch}{1.1}
  \begin{tabular}{@{}l r@{}}
    \toprule
    Model / Test subset & CER (\%) \\
    \midrule
    \multicolumn{2}{c}{\textbf{Kraken}~\cite{kiessling2019kraken,bors2024rec}} \\
    \quad Maghrebi (book\_03) & 32.71 \\
    \quad Ruq\textquoteleft ah (book\_05)   & 11.65 \\
    \quad Naskh (book\_09)    & 22.62 \\
    \quad \textbf{Overall} & \textbf{23.31} \\
    \midrule
    \multicolumn{2}{c}{\textbf{HATFormer}~\cite{chan2025hatformer}} \\
    \quad Maghrebi (book\_03) & 37.88 \\
    \quad Ruq\textquoteleft ah (book\_05)   & 13.26 \\
    \quad Naskh (book\_09)    & 25.37 \\
    \quad \textbf{Overall} & \textbf{26.74} \\
    \bottomrule
  \end{tabular}
\end{table}

Kraken's per-book CER varies substantially: 11.65\% on the
Ruq\textquoteleft ah test book, 22.62\% on the Naskh test book, and
32.71\% on the Maghrebi test book. HATFormer shows the same ordering
(13.26\%, 25.37\%, 37.88\%), so the effect is not architecture-specific.
This ordering does not track AraMS-28k training-set size by script --
Ruq\textquoteleft ah has the \emph{smallest} script-specific training
set (3,282 lines, from a single book) yet the \emph{best} test
performance, while Maghrebi has roughly 60\% more training lines
(5,229) yet the \emph{worst} test performance -- ruling out
finetuning-data volume as the primary driver. We instead attribute this
gap to script proximity in the \emph{pretrained base model}. Both Kraken and
HATFormer baselines are pretrained on Muharaf~\cite{saeed2024muharaf},
whose samples are predominantly in the Ruq\textquoteleft ah
script~\cite{saeed2024muharaf} -- a style that became the dominant
everyday hand in the Levant during the late Ottoman and early
post-Ottoman period. Our Ruq\textquoteleft ah test book therefore
benefits from a close in-distribution match with the pretraining
corpus itself, rather than from any letterform proximity to a
Naskh-targeted prior. Maghrebi, by contrast, is a paleographically
distinct tradition largely absent from Muharaf's Levantine-letters
composition -- distinct enough that it has historically required
dedicated resources rather than being folded into general Arabic HTR
corpora~\cite{vidalgorene2021rasam} -- so finetuning has to overcome a
much larger mismatch with the pretrained model's script distribution.
Naskh falls between these two poles: it is not the dominant script in
Muharaf's pretraining data, but we do not have a reported per-script
breakdown of Muharaf's composition to quantify its representation
directly, so we do not attribute the Naskh test book's intermediate
CER to a specific proximity claim. A second, non-mutually-exclusive
factor specific to book\_09 is image quality: unlike book\_05 and
book\_03, book\_09 contains a substantial number of lines with faded
or low-contrast ink (Figure~\ref{fig:script_samples}b), which
degrades recognition independently of script identity. Because each
test book is the sole representative of its script, we cannot
separate a script-level effect from this book-level covariate; the
Naskh test book's CER may therefore reflect scan/ink condition at
least as much as script distance from Muharaf's pretraining
distribution. Under this account, AraMS-28k's own Maghrebi finetuning
data is doing real work (without it, transfer from a
Ruq\textquoteleft ah-dominated pretrained model would likely be worse
still), but it is not enough on its own to close a gap that
originates upstream, in the base model's pretraining distribution. We
do not have a direct script-similarity metric (e.g., letterform
edit-distance) to quantify this claim, and leave a controlled
same-pretraining, per-script finetuning-volume ablation, together
with a verified script breakdown of the Muharaf pretraining corpus,
to future work.
\subsection{Cross-Script Generalisation Gradient}
\label{sec:gradient}
To separate the difficulty of \emph{unseen pages} from that of
\emph{unseen books} and \emph{unseen scripts}, we additionally report a
diagnostic in-distribution condition for HATFormer. For this purpose,
10\% of pages in each training book (totalling $\approx$2.5k lines)
were withheld from finetuning, which therefore used the remaining
$\approx$17.2k training lines; all HATFormer figures in this section
come from this single finetuning run. The held-out-page condition is
purely diagnostic: it is \emph{not} part of the released book-level
test split, which remains exclusively composed of fully unseen books,
and the withheld page list is distributed with the release
(\texttt{test\_seen\_pages.txt}) so the figure can be reproduced
exactly.
\begin{table}[htbp]
  \centering
  \caption{Cross-script generalisation gradient (HATFormer, line-level
  CER). The in-distribution row uses pages withheld from training
  books (a diagnostic condition outside the released test split); the
  remaining rows are the per-book results of
  Table~\ref{tab:baseline} on fully unseen books.}
  \label{tab:gradient}
  \small
  \renewcommand{\arraystretch}{1.15}
  \begin{tabular}{@{}llr@{}}
    \toprule
    Condition & Setting & CER (\%) \\
    \midrule
    In-distribution & Unseen pages, seen books & 6.48 \\
    Unseen book & Ruq\textquoteleft ah (book\_05) & 13.26 \\
    Unseen book & Naskh (book\_09) & 25.37 \\
    Unseen book & Maghrebi (book\_03) & 37.88 \\
    \bottomrule
  \end{tabular}
\end{table}
The gradient in Table~\ref{tab:gradient} is the central diagnostic
result of the benchmark. On in-distribution pages the recogniser
reaches 6.48\% CER, in the range of line-level results reported on
contemporary Arabic benchmarks. Moving to a wholly unseen
Ruq\textquoteleft ah book roughly doubles the error (13.26\%), an
unseen Naskh book roughly quadruples it (25.37\%), and an unseen
Maghrebi book -- a calligraphic tradition severely under-represented in
public training corpora -- reaches 37.88\%. This quantifies, rather
than merely asserts, the generalisation gap the dataset is built to
expose: proximity to the training distribution matters more than
aggregate finetuning volume, consistent with the script-proximity
account above. It also turns AraMS-28k's multi-script coverage from a
descriptive property into a measurable research target: closing the
in-distribution-to-Maghrebi gap is now something a later system can be
scored on. Consistent with this account, removing the intermediate
Muharaf finetuning stage raises in-distribution CER from 6.48\% to
7.44\%, confirming that exposure to a large, diverse body of real
historical Arabic before specialising on our corpus transfers usefully.
%%% ── 7  CONCLUSION ──────────────────────────────────────────────────────────
\section{Conclusion}
\label{sec:conclusion}
We presented AraMS-28k, the largest publicly released line-level
dataset of historical Arabic manuscripts, comprising 3,043 pages
spanning three hand-copied script traditions and one lithographed
printed edition, annotated with main/margin layout labels and
insertion anchors that recover the non-linear reading order of
marginal content, and released with both diacritised and
diacritic-normalised transcriptions for every line. Extended per-book
statistics and the full annotation schema are given in the appendix.
Baseline HTR experiments show the dataset is usable for finetuning
existing recognisers, and the cross-script generalisation gradient
(\cref{sec:gradient}) provides a reference point for future work on
historical Arabic manuscript recognition, layout analysis, and
reading-order recovery.
\paragraph{Limitations.} The line-validated subset is conditioned on
agreement between OCR and reference transcription (\cref{sec:qc}), which
plausibly biases it toward cleaner scans and more regular script forms;
we report it separately from the page-validated subset to make this
distinction visible. Roughly 70\% of margin lines lack a confident
insertion anchor, by design (\cref{sec:schema}): we do not force an
anchor onto content with no unambiguous attachment point in the main
text. The test set contains one book per script, so per-script CER in
\cref{sec:baselines} may partly reflect book-specific characteristics
rather than script difficulty in general. Finally, margin lines make up
a small fraction of the corpus (629 of 28,600 lines, $\approx$2.2\%);
tasks that require balanced main/margin examples, such as training a
margin-detection model from AraMS-28k alone, should account for this
imbalance, for instance via resampling or class weighting.
\paragraph{Reproducibility.} All experiments reported in
\cref{sec:baselines} use the fixed train/val/test splits distributed
with the release (Appendix~\ref{app:release}) together with the
published manifest files and checksums -- including the
\texttt{test\_seen\_pages.txt} manifest for the diagnostic
in-distribution condition of \cref{sec:gradient} -- so that results can
be reproduced exactly without re-deriving splits or re-running the
construction pipeline.
\paragraph{Data availability.} AraMS-28k is released in two
complementary formats, described in full in Appendix~\ref{app:release}:
a \textbf{full annotation release} (\textbf{AraMS-28k}) preserving every
field described in \cref{sec:schema} -- geometry, both raw and
normalised transcriptions, margin anchors, and review metadata -- and a
\textbf{recognition-ready release} (\textbf{AraMS-28k-HTR}) of segmented
line images paired with normalised ground-truth text, suitable for
direct use with standard HTR training pipelines. Both are distributed
under CC~BY-NC-SA~4.0; the RefLAM annotation pipeline used to construct
the corpus is described separately in~\cite{guechaoui2026reflam}.
%%% ── REFERENCES ─────────────────────────────────────────────────────────────
\bibliographystyle{unsrtnat}
\bibliography{references}
\FloatBarrier
%%% ── APPENDIX ────────────────────────────────────────────────────────────
\appendix
\section{Per-Book Statistics}
\label{app:perbook}
Table~\ref{tab:perbook} reports per-book statistics for both
construction phases described in \cref{sec:qc}. ``C=100'' is the
percentage of reviewed main lines that achieved a perfect alignment
score against the reference transcription; for line-validated (LV)
books this is 100\% by construction, since only perfectly aligned lines
were retained.
\begin{table*}[tbp]
  \centering
  \caption{Per-book statistics. PV = page-validated; LV = line-validated.
  The \emph{Script} column reports production format for
  \texttt{book\_10} (Lithograph), which is a lithographed printed
  edition rather than a hand-copied manuscript in one of the three
  script traditions; see \cref{sec:construction} for details.}
  \label{tab:perbook}
  \renewcommand{\arraystretch}{1.1}
  \resizebox{\textwidth}{!}{%
  \begin{tabular}{@{}lllrrrrrrll@{}}
    \toprule
    Book & Type & Script & Pages & Lines &
    Margin lines & Avg.\ words/line & Avg.\ lines/page & C=100 & Subject & Century (CE)\\
    \midrule
    \texttt{book\_03} & PV & Maghrebi     & 172 & 3,661 & 168 & 10.2 & 21.3 & 8.4\%  & Fiqh/Tafsir     & 12--13 \\
    \texttt{book\_05} & PV & Ruq\textquoteleft ah & 95  & 2,028 & 74 & 13.9 & 21.3 & 28.0\% & Hadith          & 9 \\
    \texttt{book\_06} & PV & Naskh        &  41 &   874 & 15 & 11.3 & 21.3 & 99.9\% & Qira'at/Tajwid  & 15 \\
    \texttt{book\_09} & PV & Naskh        &  46 & 1,057 &  7 & 15.5 & 23.0 & 9.8\%  & Hadith/Rijal    & 14 \\
    \texttt{book\_10} & PV & Lithograph   &  13 &   674 &  0 & 17.2 & 51.8 & 14.1\% & Medicine        & 11 \\
    \texttt{book\_11} & PV & Naskh        &  30 &   612 &  1 & 12.2 & 20.4 & 12.1\% & Aqidah          & 8 \\
    \texttt{book\_27} & PV & Naskh        & 151 & 2,532 & 364 & 13.5 & 16.8 & 11.2\% &Tasawwuf/Aqidah   & 11 \\
    \midrule
    \textit{PV subtotal} & & & 548 & 11,438 & 629 & 13.0 & 20.9 & 21.0\% & & \\
    \midrule
    \texttt{book\_12} & LV & Naskh   & 144 & 2,015 & — & 12.9 & 14.0 & 100\% & Tafsir/Qur'an   & 16 \\
    \texttt{book\_16} & LV & Naskh   & 392 & 2,276 & — & 11.2 &  5.8 & 100\% & Hadith          & 10 \\
    \texttt{book\_17} & LV & Naskh   & 584 & 3,731 & — & 12.4 &  6.4 & 100\% & Aqidah          & 11 \\
    \texttt{book\_19} & LV & Maghrebi& 122 &   666 & — & 13.6 &  5.5 & 100\% & Hadith          & 8 \\
    \texttt{book\_20} & LV & Maghrebi& 312 & 2,163 & — & 12.9 &  6.9 & 100\% & Hadith          & 8 \\
    \texttt{book\_21} & LV & Ruq\textquoteleft ah & 496 & 3,282 & — &  9.5 &  6.6 & 100\% & Hadith  & 12 \\
    \texttt{book\_24} & LV & Maghrebi& 445 & 2,400 & — & 12.8 &  5.4 & 100\% & Aqidah          & 11 \\
    \midrule
    \textit{LV subtotal} & & & 2,495 & 16,533 & — & 11.9 & 6.6 & 100\% & & \\
    \midrule
    \textbf{Total} & & Mixed & \textbf{3,043} & \textbf{27,971} & \textbf{629} & \textbf{12.3} & \textbf{9.2} & & & \\
    \bottomrule
  \end{tabular}%
  }
\end{table*}
\FloatBarrier
\section{Datasheet Summary}
\label{app:datasheet}
We summarise our documentation following Gebru et
al.~\cite{gebru2021datasheets}. \textbf{Motivation:} to provide a
layout-annotated resource for Arabic HTR and manuscript layout analysis,
addressing the margin/insertion-anchor gap identified in
\cref{sec:related}, and to show that a reference-grounded, LLM-assisted
pipeline can produce a corpus at this scale far faster than purely
manual annotation, making large-scale historical Arabic manuscript
annotation tractable. \textbf{Composition:} 3,043 page images and their
associated line-level records; the corpus depicts manuscript pages, not
individually identifiable people. \textbf{Collection:} described in
\cref{sec:construction}; page-image sources and per-book access terms
are documented in the release manifest. \textbf{Preprocessing and
labelling:} raw multimodal-LLM OCR output is never treated as ground
truth on its own; every line is grounded against an independently
sourced reference transcription and verified by a human reviewer before
release (\cref{sec:qc}). Because the reference transcription is fully
vocalised while the manuscript hand generally is not, a
diacritic-normalised transcription (\texttt{gt\_normalized}) is derived
and released alongside the raw one (\texttt{gt\_raw}) for every line
(\cref{sec:schema}). \textbf{Intended uses:} Arabic HTR, text-line
detection and segmentation (PV subset), manuscript layout analysis,
reading-order recovery, and restoration/robustness evaluation via the
flagged naturally-degraded subset (\cref{sec:dataset}). \textbf{Distribution:}
released under CC~BY-NC-SA~4.0 as described in the Conclusion.
Line-level annotations are original to this work; reference
transcriptions may carry independent copyright where derived from a
modern critical edition, and users should consult the per-source
licensing notes distributed with the release.

\subsection{Finetuning Hyperparameters}
\label{app:finetuning-details}

\paragraph{HATFormer~\cite{chan2025hatformer}.}
\begin{itemize}[leftmargin=*,noitemsep]
  \item Optimizer: AdamW ($\beta_1{=}0.9$, $\beta_2{=}0.999$,
        $\epsilon{=}10^{-6}$, weight decay 0.01)
  \item Learning rate: $5\times10^{-5}$ (half an order of magnitude
        below the from-scratch rate)
  \item Batch size: 8; warmup: 500 linear steps; gradient clipping: 5.0
  \item Training length: up to 20,000 steps, checkpointed every 1,000
        steps and selected by validation CER (greedy decoding)
\end{itemize}

\paragraph{Kraken~\cite{kiessling2019kraken}.}
\begin{itemize}[leftmargin=*,noitemsep]
  \item Initialized from the Muharaf-pretrained
        checkpoint~\cite{bors2024rec}; standard CTC training recipe
  \item Backbone frozen for epoch 1 (2,468 steps; 19,739 training
        lines, batch size 8), matched by a 2,468-step linear warmup
  \item Learning rate: $5\times10^{-4}$, cosine decay; batch size 8;
        resizing and on-the-fly augmentation enabled
  \item Early stopping: patience of 15 epochs without validation CER
        improvement, up to 100 epochs; seed 42
\end{itemize}

\section{Release Formats and Availability}
\label{app:release}
AraMS-28k is distributed as two artifacts derived from the same
underlying annotations, so that users can choose the format matching
their task without re-deriving it themselves.
\subsection{AraMS-28k (Full Annotation Release)}
The canonical release. Contains full-page images and one JSONL record
per line, following the schema in \cref{sec:schema}: \texttt{line\_type},
geometry (baseline, boundary polygon, or bounding box), both
\texttt{gt\_raw} and \texttt{gt\_normalized} transcriptions,
\texttt{gemini\_raw}, alignment confidence, margin-anchor metadata, and
review flags. This is the format to use for layout analysis,
reading-order recovery, or any task that needs geometry, diacritics, or
provenance beyond a bare image/text pair.
\begin{lstlisting}
AraMS-28k/
|-- images/{book_id}/{book_id}_{page_id}.jpg
|-- annotations/{book_id}.jsonl
|-- splits/{train,val,test}_books.txt
`-- schema/line_record.schema.json
\end{lstlisting}
Two auxiliary files are distributed alongside the main archive rather
than inside it, so they can be revised independently of the core
release: \texttt{test\_seen\_pages.txt} lists the $\approx$2.5k
lines (10\% of pages per training book) withheld from HATFormer
finetuning for the in-distribution diagnostic reported in
\cref{sec:gradient} -- a condition outside the released
book-level test split, provided purely so that diagnostic figure can
be reproduced exactly -- and \texttt{real\_damage\_lines.txt}
lists lines flagged for scan degradation, referenced in the
discussion of the Naskh test book in \cref{sec:baselines}.
\subsection{AraMS-28k-HTR (Recognition-Ready Release)}
A derived, training-ready format: one cropped line image and one
\texttt{.gt.txt} file per line, using \texttt{gt\_normalized} as the
transcription target for the reasons discussed in \cref{sec:schema}.
Directory layout follows the convention used by common CTC/seq2seq HTR
training pipelines (including Kraken), and split manifests are provided
so results are directly comparable to \cref{sec:baselines}.
\begin{lstlisting}
AraMS-28k-HTR/
|-- images/
|   |-- {book_id}_{page_id}_line{line_idx:03d}.png
|   `-- {book_id}_{page_id}_line{line_idx:03d}.gt.txt
|-- train_manifest.txt
|-- val_manifest.txt
|-- test_manifest.txt
|-- metadata.csv
|-- dataset_stats.json
`-- split_used.json
\end{lstlisting}
AraMS-28k-HTR is generated from AraMS-28k by
\texttt{scripts/build\_htr\_data.py} and is fully reproducible from the full
release; we distribute it pre-built so that no user needs to run the
segmentation/cropping step themselves.
\subsection{Hosting, Versioning, and Integrity}
\label{app:hosting}
Both releases are hosted on Hugging Face Datasets and archived on
Zenodo for long-term availability and DOI assignment. SHA-256
checksums for every archive are published alongside the release and
should be verified after download.
Table~\ref{tab:release_sizes} lists the release artifacts.
\begin{table}[htbp]
  \centering
  \caption{Release artifacts at a glance. Sizes are approximate.}
  \label{tab:release_sizes}
  \small
  \begin{tabular}{@{}lccc@{}}
    \toprule
    Release & Format & Size (approx.) & DOI \\
    \midrule
    AraMS-28k        & Images + JSONL      & 2.05 GB & \url{https://doi.org/10.5281/zenodo.22095333} \\
    AraMS-28k-HTR    & PNG crops + .gt.txt  & 2.42 GB & \url{https://doi.org/10.5281/zenodo.21499649} \\
    \bottomrule
  \end{tabular}
\end{table}
Both releases are versioned (\texttt{v1.0} at time of writing); any
future correction to transcriptions, anchors, or splits will be issued
as a new minor version with a changelog, rather than silently mutating
existing files, so that results reported against a given version remain
reproducible.
\subsection{Maintenance Plan}
The authors intend to accept correction reports (mis-transcribed lines,
incorrect anchors, or segmentation errors) via the project repository's
issue tracker. Validated corrections will be batched into versioned
releases rather than applied continuously. No further manuscript books
are currently planned for addition; extensions to new scripts or
languages, if undertaken, will be released as a distinct dataset rather
than folded into AraMS-28k.
\end{document}